\documentclass[pdflatex,sn-mathphys-ay]{sn-jnl}

\usepackage{caption}
\usepackage{subcaption}
\usepackage{enumitem}
\usepackage{adjustbox}
\usepackage{graphicx}
\usepackage{booktabs}
\usepackage{multirow}
\usepackage{amsmath,amssymb,amsfonts}
\usepackage{amsthm}
\usepackage{mathrsfs}
\usepackage[title]{appendix}
\usepackage{xcolor}
\usepackage{textcomp}
\usepackage{manyfoot}
\usepackage{algorithm}
\usepackage{algorithmicx}
\usepackage{algpseudocode}
\usepackage{listings}
\usepackage{bbding}

\begin{document}

\title[Article Title]{Modelling Adjectival Modification Effects\\ on Semantic Plausibility}


\author*[1]{\fnm{Anna} \sur{Golub}}
\equalcont{These authors contributed equally to this work.}

\author*[1]{\fnm{Beate} \sur{Zywietz}}
\equalcont{These authors contributed equally to this work.}

\author*[1]{\fnm{Annerose} \sur{Eichel}}\email{annerose.eichel@mail.de}

\affil[1]{\orgdiv{Institute for Natural Language Processing}, \orgname{University of Stuttgart}, \orgaddress{\street{Pfaffenwaldring 5b}, \city{Stuttgart}, \postcode{70569}, \state{Baden-Württemberg}, \country{Germany}}} 

\abstract{While the task of assessing the plausibility of events such as 
``news is relevant'' 
has been addressed by a growing body of work, less attention has been paid to capturing \textit{changes} in plausibility as triggered by event modification. Understanding changes in plausibility is relevant for tasks such as dialogue generation, commonsense reasoning, and hallucination detection, as it allows to correctly model,
for example, 
``\textit{false} news is relevant'', which is of lower relevance but higher concern due to potential disinformation.
In this work, we tackle the \textsc{Adept} challenge benchmark \citep{emami-etal-2021-adept} consisting of 16K English sentence pairs differing by exactly one adjectival modifier (e.g., \textit{false}.) Our modeling experiments 
provide a conceptually novel method using sentence transformers and reveal that sentence transformers struggle despite their conceptual alignment with the task at hand, underperforming in comparison to transformers like RoBERTa. 
Finally, we discuss our findings in relation to prior work and present a detailed error analysis to shed light on potential sources for ST underperformance, highlighting advantages
and shortcomings of the examined methods for balancing out train and test data. 
}




\maketitle

\section{Introduction}

Capturing and modeling change in semantic plausibility is a crucial building block when processing natural language. 
The challenge of estimating the plausibility of events such as
``news is relevant''  or ``sarcasm brings hostility'' 
has been addressed by a growing body of work from various perspectives
\citep{van-de-cruys-2014-neural,wang-etal-2018-modeling,porada-etal-2019-gorilla,pedinotti-etal-2021-cat,liu2023vera,eichel-schulte-im-walde-2024-multi,kauf2023event,kauf2024comparingplausibilityestimatesbase}
but less attention has been devoted to exploring changes in plausibility triggered by event modification. Consider, for example, a change upon inserting the adjectival modifier `false' into the example above, now referring to considerably less relevant ``\textit{false} news'', or inserting `gentle' to form ``\textit{gentle} sarcasm'', now referring to a sign of closeness rather than unkindness
\citep{jang2023intended}.
Exploring and understanding models' abilities to handle such modifications and corresponding changes in event plausibility is relevant for downstream tasks where capturing variation in event plausibility is crucial, such as 
dialogue generation \citep{ko-etal-2019-linguistically}, 
commonsense reasoning, explanation \citep{wang2020semeval2020}, and verification \citep{liu2023vera}, 
natural language inference \citep{li-etal-2019-learning}, information retrieval \citep{eichel-etal-2023-made}, 
and hallucination detection \citep{mickus-etal-2024-semeval}.

Notably, \citet{emami-etal-2021-adept} introduce the benchmark dataset \textsc{Adept} consisting of 16K English sentence pairs differing by one adjectival modifier, in order to examine the ability of language models (LMs) to capture event plausibility. Evaluating a range of models, they find that LMs struggle to reach performance on par with human judges and highlight the LMs' lack of attention to context and commonsense reasoning.

In this study, we present a novel, conceptually different modeling approach to tackle the \textsc{Adept} challenge task, drawing on sentence transformers (ST) \citep{reimers-gurevych-2019-sentence}. 
ST take two input sequences. Using siamese and triplet network structures, semantically meaningful sentence embeddings can be derived and compared drawing on metrics such as cosine similarity.
This has led to improved performance in areas such as semantic similarity \citep{reimers-gurevych-2019-sentence} and general and domain-specific sentence classification \citep{smedbert,Alfianto2023}.
We conduct an in-depth analysis of the challenge dataset, develop a new plausibility model and propose a more realistic evaluation approach.
Our results are accompanied by an analysis of potential sources for ST underperformance regarding the task at hand.
We further discuss relevant aspects of \citet{emami-etal-2021-adept}'s and our work including model performance, evaluation approach, and model error analysis. 

Specifically, we (i)~replicate \citet{emami-etal-2021-adept}'s  \textbf{transformer}-based modeling approach using BERT \citep{devlin-etal-2019-bert}, RoBERTa \citep{roberta}, and DeBERTa \citep{deberta} on an adapted \textsc{Adept} dataset, and (ii)~assess \textbf{sentence transformers}, a conceptually different model architecture proven successful for modeling sentence similarity and classification \citep{reimers-gurevych-2019-sentence,smedbert,Alfianto2023}
on the task of predicting change in plausibility. Our findings can be summarized as follows:

\begin{enumerate}[label=(\roman*)]
    \item \textbf{Transformer}-based models and \textbf{sentence transformers}
    \textbf{struggle with predicting plausibility changes} triggered by an adjectival modifier. 
    \item Although a better conceptual fit, \textbf{sentence transformers underperform} classical \textbf{transformer}-based approaches
    across the considered experimental setups with a fine-tuned RoBERTa model emerging as the best-performing option.
    \item We highlight the relevance of \textbf{training data balance} for \textbf{model performance} as well as \textbf{test data balance} for \textbf{evaluation result reliability}. 
\end{enumerate} 

The accompanying code is publicly available.\footnote{\url{https://github.com/RedLuckyPanda/Adjectival-Modification-Semantic-Plausibility}}
\section{Related Work}

\subsection{Capturing Changes in (Semantic) Plausibility}

The notion of plausibility has been explored from various perspectives, including \textit{selectional preferences} \citep{erk-etal-2010-flexible,van-de-cruys-2014-neural,metheniti-etal-2020-relevant}
and \textit{thematic fit} \citep{chersoni-etal-2016-towards,sayeed-etal-2016-thematic,pedinotti-etal-2021-cat} where plausibility estimations are understood to reflect non-surprisal in a given context.
However, in the context of (semantic) plausibility, plausible events are not necessarily the most typical or preferred. This contrasts with selectional preference or thematic fit, where non-preferred events are atypical albeit still plausible. 
\citet{wilks1975} notes that the most preferred option might not always lead to the only correct interpretation and emphasizes the need to ``prefer the normal, but \textit{accept} the unusual.'' To this end,
\citet{wang-etal-2018-modeling} formulate the task of semantic plausibility as ``recognizing plausible but possibly novel events''. In the context of binary or multi-class semantic plausibility classification, the goal is thus to develop models to distinguish between plausible and implausible events or sequences \citep{wang-etal-2018-modeling,porada-etal-2019-gorilla,porada-etal-2021-modeling,bang-etal-2023-multitask,eichel-schulte-im-walde-2024-multi}.
Further previous related work providing insights from various angles includes \citet{liu2023vera} who propose a commonsense plausibility model to validate declarative natural language statements as well as \citet{tang2023less} focusing on the generation of relevant plausible but ``less likely'' alternative hypotheses for a given statement in a commonsense reasoning task. 

Providing insights from a different perspective including NLI and RTE,
\citet{emami-etal-2021-adept} introduce the challenge task of capturing \textit{change} in plausibility to ``understand the strengths and weaknesses of LMs that have led to state-of-the-art performance in downstream NLI-tasks''. They set the task of classifying sentence pairs differing by one adjectival modifier to ``test a system’s ability to correctly interpret and reason about adjective-noun pairs within a given context''. To do so, \citet{emami-etal-2021-adept} devise the \textsc{Adept} dataset 
encompassing 16K English sentence pairs annotated with crowd-sourced human ratings. More specifically, human annotators assess whether the insertion of an adjective leads to a modification of the semantic plausibility of a sentence. Furthermore, \citet{emami-etal-2021-adept}  fine-tune and evaluate a range of transformer-based models in in-context and no-context settings, as well as a rule-based approach taking into account adjective taxonomies. To compare models to human performance, a sample of 100 in-context and no-context dev set instances is annotated by one human annotator and evaluated wrt. agreement to crowd-sourced majority labels.
Results indicate a substantial gap between human and model performance,
stagnating at a dev set\footnote{We have contacted the authors who confirmed that they (i) used the \textit{dev} set for model optimization and (ii) evaluated and reported \textit{dev} set results only.}  accuracy of 0.71 (vs. human performance of 0.85). While human performance drops significantly in case of a no-context setting, model performance is basically on par with the full context setting, suggesting low model vs. high human sensitivity to context. Furthermore, the used adjective classes do not provide sufficient information to correctly predict changes in plausibility.
In contrast, our work considers change in plausibility from a novel perspective. We analyze, adapt, and leverage the \textsc{Adept} dataset to train and evaluate a range of semantic plausibility models, providing insights into dataset quality for the challenge task at hand and a comparison of evaluation approaches taking into account both replicated and newly introduced models.

\subsection{Modeling Semantic (Dis)similarity}
Modeling similarity between inputs has been approached using, among many other models and methods, Siamese encoder models (SE), which concurrently process two inputs and map them on a single scalar output. Sentence transformers (ST) present one SE model realization which learns to predict a similarity score between two input sequences. STs have lead to statistically significant improvements in many areas including semantic similarity \citep{reimers-gurevych-2019-sentence}, in- and out-of-domain information retrieval \citep{thakur-etal-2021-augmented}, 
as well as general and domain-specific sentence classification \citep{smedbert}.
To our knowledge, however, ST models have not been fine-tuned and evaluated for the task of predicting similarity or changes between sentence pairs differing in an adjectival modifier to elicit a semantic plausibility judgment. 

\section{Data}
\label{sec:data}

\subsection{The \textsc{Adept} Challenge Dataset}

The \textsc{Adept} dataset \citep{emami-etal-2021-adept} consists of 16,115 English sentence pairs differing only in an adjective modifying a noun, e.g., \{A horse goes away $\leftrightarrow$ A \textit{dead} horse goes away\}. 
To create \textsc{Adept}, \citet{emami-etal-2021-adept} scrape English Wikipedia and Common Crawl samples for high-frequency adjectival modifier-noun pairs. 
After filtering the data and adapting adjective distributions, sentences are constructed using relevant predicates for the noun in an adjectival modifier-noun pair found through ConceptNet \citep{speer2017conceptnet}. The extracted adjectival modifiers are inserted in the constructed sentences, generating four variants per sentence.
Each sentence set is annotated through crowd-sourcing with one of the labels: \{\textit{impossible, less likely, equally likely, more likely, necessarily true}\}. 
The labels are defined to capture whether and how sentence plausibility changes upon the insertion of an adjectival modifier.
The label \textit{\textbf{impossible}} captures the extreme cases of a shift from a plausible event to an improbable or illogical incident, e.g. 
\{a horse goes away $\leftrightarrow$ a \textit{dead} horse goes away\}, while \textit{\textbf{necessarily true}} signifies the option of turning a plausible but not necessarily true statement into exactly that, e.g., \{a statement is a lie $\leftrightarrow$ a \textit{false} statement is a lie\}. 
In turn, the labels \textit{\textbf{less likely, equally likely, more likely}} are designed to capture whether sentence plausibility gradually changes \textit{in comparison to the previous sentence}
without necessarily including information regarding the real-world plausibility of the sentence itself.
Quality control measures include discarding submissions from workers failing attention and control checks against gold annotations, limitations on the maximum number of submitted instances as well as excluding sentence pairs where less than two annotators agree on a final label.
Importantly, despite multiple quality control measures, an analysis of final annotations reveals higher disagreement for more extreme labels and an emphasis on modified sentence plausibility only as compared to mid-range labels. Annotators do agree, however, more often than not on the \textit{directionality} but not the \textit{degree} of change in plausibility with the source of disagreement more often a different label on the same side of the scale, e.g., \textit{less likely} and \textit{impossible} than on two opposing sides, e.g., \textit{less likely} and \textit{more likely}.  

\begingroup
\setlength{\tabcolsep}{9pt} 

\begin{table}[!t]
\caption{Label distribution of the original and adapted \textsc{Adept} train sets. 
Classes 0--4 represent \{\textit{impossible, less likely, equally likely, more likely, necessarily true}\}.
}

\small
\centering
\begin{tabular}{@{}lrrrrr@{}}
\toprule
         & 0 & 1 & 2 & 3 & 4 \\ \midrule
\citet{emami-etal-2021-adept} & 0.14       & 0.12        & 0.67           & 0.07        & 0.01             \\ \midrule
\textsc{full}     & --       & 0.14        & 0.77           & 0.09        & --          \\
\textsc{balanced}  & --       & 0.38        & 0.38           & 0.24        & --   \\
\bottomrule
\end{tabular}
\label{tab:dataset-stats}
\vspace{-0.4cm}
\end{table}

\endgroup

\subsection{Adapting \textsc{Adept}} We focus on \citet{emami-etal-2021-adept}'s mid-range classes, thus excluding the labels \textit{impossible} and \textit{necessarily true}.
We base this decision on \citet{emami-etal-2021-adept}'s analysis of the annotators' use of these labels, which found that annotators could assign these meaningfully in only up to one third of the cases.
Instances of the label \textit{necessarily true}
make up only 1\% of the dataset but with lowest IAA, while approx. 50\% of instances labeled \textit{impossible} are false positives, ungrammatical, or nonsensical.
Meanwhile, modeling change in plausibility requires the sentences to be well-formed \textit{and} plausible, making this dataset part unfit for the task at hand.\footnote{This is in line with \citet{emami-etal-2021-adept} who also consider adapted versions of their initial 5-class setting. They, however, merge extreme labels with other labels or discard the label \textit{impossible} only.} 
In addition, we address the dataset's severe class imbalance present in both original and adapted dataset setups by down-sampling the class \textit{equally likely} to a total of 1,500 sentence pairs, matching the size of the class \textit{less likely} and approaching that of \textit{more likely} (see Table~\ref{tab:dataset-stats}).\footnote{\citet{emami-etal-2021-adept} release majority-aggregated labels which do not include annotation decisions by individual annotators. Thus, it is not possible to assess whether our adapted three-class subset shows systematic differences in annotation quality or IAA relative to the full dataset.} 

\section{Models}

\subsection{Majority Prediction}
The majority baseline always predicts the label \textit{equally likely} when fine-tuning on the \textsc{full}, and \textit{less likely} when using the \textsc{balanced} train set.\footnote{While Table~\ref{tab:dataset-stats} indicates equal label distribution for both equally likely and less likely, the latter technically outnumbers the former by 3 instances.}

\subsection{Transformers} \label{subsec:transformers}
To model semantic plausibility, we fine-tune \texttt{BERT-base-uncased} \citep{devlin-etal-2019-bert}, \texttt{DeBERTa-base} \citep{deberta}, \texttt{RoBERTa-base} \citep{roberta} and \texttt{MPNet-base} \citep{mpnet}.  For details on used model implementations, cf. App.~\ref{appsec:exp}.

We train each model to predict class labels using (i) full imbalanced (\textsc{full}) and (ii) balanced (\textsc{balanced}) train data to test the effect of class balancing. The models are fed tokenized sentence pairs with the input formatted as \textit{[CLS] sentence1 [SEP] sentence2 [SEP]}.
The target labels are 0 for \textit{less likely}, 1 for \textit{equally likely}, and 2 for \textit{more likely}. Each transformer model is fine-tuned for 3 epochs\footnote{In comparison to prior \citep{emami-etal-2021-adept} and related \citep{pyatkin-etal-2021-possible,anthonio-etal-2022-clarifying} work, the number of training epochs is on the lower end. This is due to the fact that we work with less available training data, especially when reducing class imbalance (\textsc{Balanced}) which leads to models overfitting after 2-3 epochs. We address this by evaluating models trained for 3 epochs as a compromise between learning more from the train set (i.e. lower train loss) and retaining the ability to generalize to unseen data (i.e. unchanged val loss).}, using a batch size of 16, and an Adam optimizer with default hyperparameters. The learning rate starts at 2e-5 and decreases linearly as training progresses.


\subsection{Sentence Transformers (ST)} \label{subsec:st}
In contrast to transformer models such as BERT that are not originally built for comparing two sentences, ST models process two sequences in parallel. 
In the context of predicting change in semantic plausibility, we explore whether ST advantages can be successfully leveraged by constructing separate embeddings for modified and unmodified sentences. Next, the embeddings are fine-tuned to capture and express the differences between the (un-)modified sentences. 
For our experiments, we use the general purpose ST model \texttt{all-mpnet-base-v2} based on MPNet \citep{mpnet}, which is pretrained on a semantic search objective using more than 1B sentence pairs \citep{reimers-gurevych-2019-sentence}. For details on the used model implementation, cf. App.~\ref{appsec:exp}.
To fine-tune the model, we structure our input as follows: Each instance is represented as a list containing the first sentence \underline{without} the modifier, the second sentence \underline{with} the modifier, and the corresponding label. We fine-tune ST with the MPNet backbone using standard hyperparameters and a learning rate of 2e-5, while cosine similarity serves to compare sentence embeddings and to extract a label. As the target scores need to be valid cosine similarities, we map our three classes to the values 0, 0.5 and 1. We train the models for 4 epochs.\footnote{The choice of the number of epochs is motivated similarly to §\ref{subsec:transformers}.}

To evaluate our models, we 
divide the continuous cosine similarity range into classes,
matching scores $\leq0.33$ to \textit{less likely}, 
$0.34 - 0.65$ 
to \textit{equally likely}, and 
$\geq0.66$ 
to \textit{more likely}.
Thresholds and label mappings are determined empirically following experimentation with different values. As the embedding space is adapted to the chosen labels during training, different threshold values lead to similar results so long as they split the value range defined by the label mappings into three equal parts. More specifically, our experiments on varying the thresholds show that thresholds based on a vanilla ST results in values that are too close together to allow for meaningful differentiation of classes ($\leq$0.87 and  $\geq$0.89). Thresholds that define a larger range dedicated to capturing the middle class \textit{equally likely} substantially increase recall; however, the resulting loss in precision wrt. the other classes renders this setting unsuitable. Furthermore, different label-mappings, such as assigning labels to $-$1, 0, and 1 to use the full range of possible cosine similarities, similarly do not improve results. Based on these considerations, threshold values are set to 0.33 and 0.66, which partition the considered interval [0,1] into three equal parts.\footnote{Class-wise average similarity scores (class 1: 0.12, class 2: 0.52, class 3: 0.82) support this decision. Considering the average similarity scores assigned to instances of each class shows that the median between the similarities (0.32 and 0.67) is similar to the chosen 0.33 and 0.66 threshold values. However, using the exact median values as threshold values is expected to reduce generalization on new data, which is why even partitioning was used.}

\section{Metrics and Evaluation}
\label{subsec:eval}
We evaluate all our models on the test set reporting macro F1 and accuracy averaged over 3 runs. 
In addition, we report ROC-AUC scores for all models except ST models. This is due to the fact that in our setup, the model output consists of single cosine similarity score per data point.\footnote{While we do not report ROC-AUC scores for ST, ROC-based evaluation is technically possible from a single similarity score, for example, by varying decision thresholds to compute 
ROC-AUC scores.}

\subsection{Cross-balanced evaluation} \label{subsec:cbe}
\begin{figure}[!t]
    \centering
    \includegraphics[width=\linewidth]{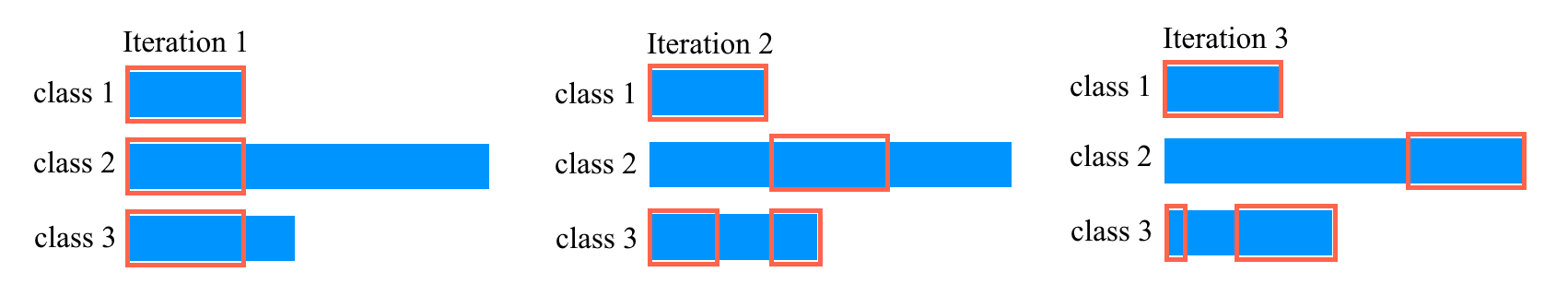}
    \caption{Three iterations of \textit{\textbf{cross-balanced evaluation}}. The window (red) is shifted across dataset classes until every instance has been seen (not to scale)
    }
    \label{fig:1 - cross_bal_eval}
    \vspace{-0.4cm}
\end{figure}

While we address dataset imbalance during training through majority-class down-sampling, modifying evaluation sets would affect validity and comparability with other models tested on the same data. At the same time, evaluating on severely imbalanced data might bias scores toward the majority class. To mitigate this, we propose a \textbf{cross-balanced (CB) evaluation} approach. Rather than evaluating on the full imbalanced set at once, we instead evaluate on a series of balanced subsets and average a metric across them. Each subset contains exactly $n_\text{min}$ instances from every class, where $n_\text{min}$ is the size of the smallest class. Formally, let $\mathcal{D}$ be the test set over $K$ classes, with $\mathcal{D}_k \subseteq \mathcal{D}$ denoting the instances of class $k$ and $n_k = |\mathcal{D}_k|$. We define $n_\text{min} = \min(n_{1},n_{2},n_{3})$ and
$n_\text{max} = \max(n_{1},n_{2},n_{3})$. The number of evaluation rounds is $T = \lceil n_\text{max} / n_\text{min} \rceil$.
In each round $t \in \{1, \ldots, T\}$, we form a balanced subset by taking a window of $n_\text{min}$ consecutive instances from each $\mathcal{D}_k$, cycling back to the start of $\mathcal{D}_k$ when its end is reached. Let $\mathcal{B}_k^{(t)}$ denote this window for class $k$ in round $t$ (see Fig.~\ref{fig:1 - cross_bal_eval}). The final CB score is the average of metric $M$ over all $T$ rounds: \begin{equation} \text{CB}(M) = \frac{1}{T} \sum_{t=1}^{T} M\!\left(\bigcup_{k=1}^{K} \mathcal{B}_k^{(t)}\right) \end{equation}

\smallskip

In comparison with metrics such as \textit{macro F1}, cross-balanced evaluation provides insights into different aspects of model performance. As defined in Eq.~\ref{eq:macrof1}, macro F1 first calculates class-wise F1 scores, which are then averaged
\begin{equation} \label{eq:macrof1}
F1_{\text{macro}} = \frac{1}{K} \sum_{k=1}^{K} F1_k
\end{equation}
where $K$ denotes the number of classes and $F1$ refers to the harmonic mean of precision and recall. The main difference between macro F1 as defined in Eq.~\ref{eq:macrof1} and macro F1 computed based on cross-balanced evaluation is that, in case of the latter, F1 is computed on a balanced windows of size per class. The results are then averaged over evaluation iterations. This means that the final score reflects average model performance over systematically balanced subsets of the test data, rather than a single aggregate over the full test set. 
We further compare cross-balanced evaluation to \textit{weighted F1}. In contrast to macro F1 using the arithmetic mean for aggregation, weighted F1 averages all per-class F1 scores while considering each class’s support, i.e., the number of actual occurrences of the class in the dataset. This way, original class imbalance is taken into account, which diverges conceptually from the proposed cross-balanced approach.

Considering the resampling aspect further draws attention to \textit{k}\textit{-fold cross validation} ($k$-fold CV). However, the two approaches differ in several aspects. Firstly, $k$-fold CV is a training technique, while cross-balanced evaluation 
is applied to a fixed test set and a fixed model. 
Secondly, the $k$ in $k$-fold CV can be freely determined, while the total number of evaluation rounds $T$ in cross-balanced evaluation is determined by the size of the largest and smallest considered classes. Finally, $k$-fold CV does not enforce class balance across folds. In contrast, the cycling window in our cross-balanced approach specifically accounts for class balance in every evaluation round $t$.

\subsection{Standard evaluation} For completeness and full comparability with previous work, we also evaluate our models the traditional way, i.e. on a single-run through the dev or test set. 

\subsection{Statistical significance testing} 
To ensure the statistical significance of the results, we fine-tune each model with the seeds \{6, 17, 42\} and report averaged metrics. We determine the model ranking  using \textit{Almost Stochastic Order} testing \citep{ulmer2022deep}.
For this, we compare corresponding pairs of models using ASO with a confidence level of $\alpha = 0.05$ before adjusting for all pair-wise comparisons using the Bonferroni correction. 
We consider model A to perform better than model B wrt. a metric if A is almost stochastically dominant over B 
($\epsilon_\text{min} < \tau$ with $\tau = 0.4$) 
and B is not almost stochastically dominant over A. If for a pair of models this condition is not fulfilled, we consider the difference between their performances insignificant.

\section{Results}

Table~\ref{tab:main-results} presents an overview of the transformer and sentence transformer model performance fine-tuned on \textsc{full} and \textsc{balanced} training data and evaluated on the test set using \textit{cross-balanced} and \textit{standard} evaluation. 

\paragraph{Cross-balanced evaluation} Concentrating on \textit{cross-balanced} evaluation results, we find \textbf{RoBERTa} fine-tuned on the \textsc{\textbf{balanced}} training data outperforming all other approaches at a statistically significant level, with macro F1 and ROC-AUC scores of 0.61 and 0.78, respectively. While \textbf{MPNet} is on par regarding macro F1, its ROC-AUC score is the lowest across all models. This might suggest well-separated model outputs at a chosen threshold or good performance for a smaller class, but suboptimal performance regarding ranking probabilities overall.
DeBERTa and BERT trained on \textsc{balanced} data achieve slightly lower macro F1 and ROC-AUC scores of 0.60. Based on macro F1 performance, the ST approach is ranked last with a score of 0.58.

In contrast, models fine-tuned using unbalanced \textsc{full} data perform considerably worse while still outperforming the majority prediction baseline.\footnote{Due to the cross-balanced evaluation approach described in \ref{subsec:cbe}, a random baseline would achieve a 0.34 macro F1 score.} 
Here, the best-performing model is our ST-based approach \textbf{ST-\textsc{full}} that achieves an macro F1 of 0.56, making the most of the available information in the train data. 

\begin{table*}[!t]
\vspace{-2ex}
\caption{\textit{Cross-balanced} and \textit{standard} evaluation results across baseline, transformer, and ST models fine-tuned on \textsc{balanced} (\textsc{bal}) and \textsc{full} train sets.  
The highest scores are marked in bold. 
In case of multiple highlighted values per column, the difference between corresponding models is not statistically significant, i.e., all models are considered best wrt. a metric. As described in §\ref{subsec:st}, ROC-AUC is not reported for ST models.
}

\centering
\begin{adjustbox}{width=\textwidth}
\begin{tabular}{@{}llrrrrrr@{}}
\toprule
\multicolumn{1}{c}{\textsc{model}}  & & \multicolumn{3}{c}{\textit{cross-balanced}}                             & \multicolumn{3}{c}{\textit{standard}}                                 \\
\midrule
& \multicolumn{1}{l|}{\textbf{train}} & \multicolumn{1}{l}{\textbf{macro F1}} & \multicolumn{1}{l}{\textbf{ROC-AUC}} & \multicolumn{1}{l}{\textbf{Acc}} & \multicolumn{1}{|r}{\textbf{macro F1}} & \multicolumn{1}{l}{\textbf{ROC-AUC}} & \multicolumn{1}{l}{\textbf{Acc}} \\ \midrule

Majority Pred.           & \multicolumn{1}{l|}{\textsc{bal}}           & 0.167 & - & 0.333 & \multicolumn{1}{|r} {0.045} & - & 0.073 \\ 

BERT           & \multicolumn{1}{l|}{\textsc{bal}}            & 0.596 & 0.768 & 0.597 & \multicolumn{1}{|r}{0.467} & 0.770  & 0.582 \\
                                
DeBERTa        & \multicolumn{1}{l|}{\textsc{bal}}            & 0.597 & 0.774 & 0.595 & \multicolumn{1}{|r}{0.464} & 0.771 & 0.577 \\
                                
MPNet          & \multicolumn{1}{l|}{\textsc{bal}}            & \textbf{0.612} & 0.763 & \textbf{0.612} & \multicolumn{1}{|r}{0.473} & 0.759 & 0.583 \\

RoBERTa        & \multicolumn{1}{l|}{\textsc{bal}}            & \textbf{0.612} & \textbf{0.784} & \textbf{0.612} & \multicolumn{1}{|r}{0.471} & 0.784 & 0.579 \\ 
ST            & \multicolumn{1}{l|}{\textsc{bal}}            & 0.588 & -     & 0.592 & \multicolumn{1}{|r}{0.410 } & -     & 0.475\\

\midrule
Majority Pred.           & \multicolumn{1}{l|}{\textsc{full}}           & 0.167 & - & 0.333 & \multicolumn{1}{|r}{0.296} & - & 0.798 \\  
BERT          & \multicolumn{1}{l|}{\textsc{full}}               & 0.500   & 0.742 & 0.522& \multicolumn{1}{|r}{0.537} & 0.740  & 0.781 \\
                                
DeBERTa        & \multicolumn{1}{l|}{\textsc{full}}               & 0.529 & 0.767 & 0.545 & \multicolumn{1}{|r}{0.555} & 0.765 & 0.785 \\

MPNet          & \multicolumn{1}{l|}{\textsc{full}}               & 0.510  & 0.748 & 0.532& \multicolumn{1}{|r}{0.551} & 0.744 & 0.794 \\

RoBERTa        & \multicolumn{1}{l|}{\textsc{full}}               & 0.521 & 0.775 & 0.540  & \multicolumn{1}{|r}{0.550}  & 0.773 & 0.788\\ 

ST            & \multicolumn{1}{l|}{\textsc{full}}                & 0.562 & -     & 0.564 & \multicolumn{1}{|r}{0.528} & -     & 0.726 \\      

 \bottomrule
\end{tabular}
\end{adjustbox}
\label{tab:main-results}
\end{table*}

\paragraph{Standard evaluation} Evaluating models with a \textit{standard} evaluation approach shows that \textsc{balanced} training data yields ROC-AUC scores comparable to cross-balanced evaluation (not considering ST). However, macro F1 scores are substantially lower. 
Considering both macro F1 and ROC-AUC, RoBERTa again outperforms all other models. An inspection of ST performance based on macro F1 finds the model ranked last again, except for the majority prediction baseline.

For models fine-tuned on the \textsc{full} training set, a slight decrease in ROC-AUC is observed. 
In contrast, substantially improved macro F1 scores for all models are found. This includes ST which clearly benefits from having seen the full dataset instead of a balanced version. Overall, based on macro F1, all models assessed using standard evaluation profit from seeing the full dataset, while models do not benefit from the full dataset when evaluated using cross-balanced evaluation.

\smallskip

In summary, our findings highlight (i) the importance and effect of class balance in both training and test data, (ii) the relevance of using larger and more advanced architectures such as RoBERTa over BERT, (iii) that ST, albeit seemingly the most conceptually fitting model for the task, is not robust enough to outperform transformer architectures. 

\section{Error Analysis and Discussion}

\subsection{ST Model Performance Analysis}
In the following, we examine ST underperformance regarding the investigated task of predicting change in plausibility introduced through an adjectival modifier. To do so, we first investigate whether ST models failure is connected to specific adjectival semantic categories. Second, we explore whether there are informative differences in performance across plausibility labels.

\paragraph{Semantic Categories of Adjectives} 
We analyze model predictions by ST models trained on \textsc{full} or \textsc{balanced} data. We first focus on the adjectival modifiers as they constitute the main trigger for change in plausibility in the task at hand. We consider three semantic categories, including whether an adjective (i) is relational (e.g. \textit{national}) or descriptive (e.g. \textit{fresh}), (ii) is evaluative, i.e. carries positive (e.g. \textit{free}) or negative (e.g. \textit{abandoned}) sentiment, or is rather rather neutral (e.g. \textit{standard}); (iii) denotes a color (e.g. \textit{yellow, grayish, crimson}). 

To determine whether an adjective is \textit{relational}, we look up the adjective in \texttt{WordNet} \citep{fellbaum1998}, retrieve its primary sense, and store the lexical category (\texttt{adj.all}: descriptive, \texttt{adj.pert}: pertainyms, i.e., relational). To assess whether an adjective is \textit{evaluative}, we employ \texttt{VADER} \citep{hutto2014vader} through \texttt{NLTK} \citep{nltk} to obtain polarity scores which provide information regarding sentiment. We use the \texttt{compound} score, and assign a neutral label in case  $score=0$, a negative label in case $0>score$, and a positive label in case $0<score$.\footnote{We also experiment with taking the maximum assigned polarity score instead of the compound score which does not change results.} To determine whether an adjective denotes a color, we again employ \texttt{WordNet} and find the adjective's synsets, derive its related noun forms (e.g., \textit{red} → \textit{redness}), and check if any of those nouns have a direct hypernym that is a color-related concept.

\begin{table}[!t]
\centering
\caption{Analysis of ST model errors regarding semantic categories of adjectival modifiers, with scores denoting respective shares in percent. Models are either trained on the \textsc{balanced} or \textsc{full} dataset. We analyze predictions on the adapted \textsc{Adept} test set. For comparison, we show statistics including errors and correctly predicted instances wrt. the analyzed semantic categories in italics.}
\begin{tabular}{l|rr|rr|rr}
\toprule
\textbf{ST model errors }          & relational & descriptive & evaluative & non-evaluative & color & non-color \\ \midrule
\textsc{train: balanced}         & 6.63       & 93.37       & 21.86      & 78.14          & 2.26  & 97.74    \\
\textsc{train: full}             & 6.25       & 93.75       & 21.09      & 78.91          & 1.04  & 98.96    \\
\midrule
\textit{errors + correct} & \textit{7.11}       & \textit{92.89}      & \textit{19.49}     & \textit{80.51}          & \textit{1.76}  & \textit{98.24}   \\
\bottomrule
\end{tabular}
\label{tab:ST-adjectives}
\end{table}

In Table~\ref{tab:ST-adjectives}, we present semantic category statistics for cases where the ST models fail to classify change in plausibility correctly. Models are either fine-tuned on the \textsc{balanced} or \textsc{full} dataset, using the random seed 17. We analyze predictions on the adapted \textsc{Adept} test set and add overall test set statistics for comparison. The results show that ST models do not fail in a systematic way for one of the examined semantic categories. Thus, the investigated semantic categories cannot conclusively explain ST underperformance regarding the task at hand.

\paragraph{Impact of Plausibility Class} Thus, we further explore whether there are informative differences in performance across plausibility labels. In analogy to the previous analysis, we consider ST models that are either fine-tuned  on the \textsc{balanced} or \textsc{full} dataset, using the random seed 17. We focus on predictions on the test set, and analyze whether prediction errors are independent of the true label using a $\chi^{2}$-square test of independence. We find statistically significant associations between true label and prediction error for both model variants: (i) \textsc{balanced} train data: $\chi^{2}=$36.68, $p<0.001$, (df$=$2, N$=$1380), and (ii) \textsc{full} train data: $\chi^{2}=$153.64, $p<0.001$, (df$=$2, N$=$1380). This suggest that model performance depends on plausibility labels. Hence, we conduct an inspection of class-wise error rates, i.e. the proportion of misclassified instances per class, reported in percent. Analyzing predictions by the ST model using \textsc{full} train data reveals significantly lower performance for labels denoting change in plausibility (\textit{less likely}: 0.61\%, \textit{more likely}: 0.52\%). In comparison, stability in plausibility is often predicted correctly (\textit{equally likely}: 0.21\%), which reflects the underlying unbalanced train data. Looking at results for ST model using \textsc{balanced} train data reveals error rates that mirror the attempt to balance out training data. While change in plausibility (\textit{less likely}: 0.38\%, \textit{more likely}: 0.33\%) is captured more reliably, predicted instances are more frequently misclassified where plausibility is not modified (\textit{equally likely}: 0.56\%). These observations suggest that ST models do neither profit from using the full dataset nor benefit from rigid down-sampling of majority classes to address class imbalance. Rather, a more balanced approach to resolving class imbalance or training approaches such as using weighted loss functions might present viable alternatives for future work.

\begin{table}[!t]
\caption{
Comparison of results to \citet{emami-etal-2021-adept}
across multi-class setups and evaluation approaches. Acc. denotes \textit{accuracy}.
}
\centering
\small

\begin{tabular}{l l l r}
\toprule
\textbf{Model} & \textbf{Classes} & \textbf{Evaluation} & \textbf{Acc.(Dev)} \\
\midrule
3-class DeBERTa \citep{emami-etal-2021-adept} & 0 + 1, 2, 3 + 4 & standard & 0.739 \\
4-class DeBERTa \citep{emami-etal-2021-adept} & 1, 2, 3, 4 & standard & 0.812 \\
\midrule
RoBERTa-\textsc{full} & 1, 2, 3 & standard & 0.797 \\
RoBERTa-\textsc{balanced} & 1, 2, 3 & cross-balanced & 0.612 \\
\bottomrule
\end{tabular}
\label{tab:emami-vs-us}
\end{table}

\subsection{Model Accuracy}
Comparing our results with \citet{emami-etal-2021-adept} requires both care and caution. First, the class setups differ. As presented in (Table~\ref{tab:emami-vs-us}), they either merged \textit{necessarily true} and \textit{impossible} into \textit{more likely} and \textit{less likely} for a three-class setup, or excluded only the \textit{impossible} class and used four classes. In comparison, we excluded both classes. In addition, performance metrics on both sides are reported on the dev set, which was also used for model optimization.\footnote{For clarity, we reiterate that we have contacted the authors to confirm that they (i) used the \textit{dev} set for model optimization and (ii) evaluated and reported \textit{dev} set results only.} Furthermore, accuracy is the only available metric for comparison. However, as the majority prediction baseline of \citet{emami-etal-2021-adept} illustrates, accuracy can be deceptive on \textsc{Adept}, which is heavily skewed towards \textit{equally likely}: a model predicting only that class will still yield high scores.
With these considerations in mind, our RoBERTa-\textsc{full} model evaluated using \textit{standard} evaluation achieves a similar accuracy to both the 3-class and 4-class DeBERTa models of \citet{emami-etal-2021-adept}. Their marginally higher score may reflect our choice to exclude the two problematic classes rather than merge them. Under \textit{cross-balancing} evaluation, RoBERTa-\textsc{balanced} scores lower than the 3-class DeBERTa. This is expected as cross-balancing removes the advantage that models gain from over-predicting the majority class, making it a stricter evaluation than the standard accuracy reported in prior work. However, the differences in setup and the limitations of accuracy as a metric on imbalanced data mean that these findings should be interpreted cautiously.

\begin{figure}[!t]
    \centering
    \begin{subfigure}[!htpb]{0.45\textwidth}
    \includegraphics[width=1\linewidth]{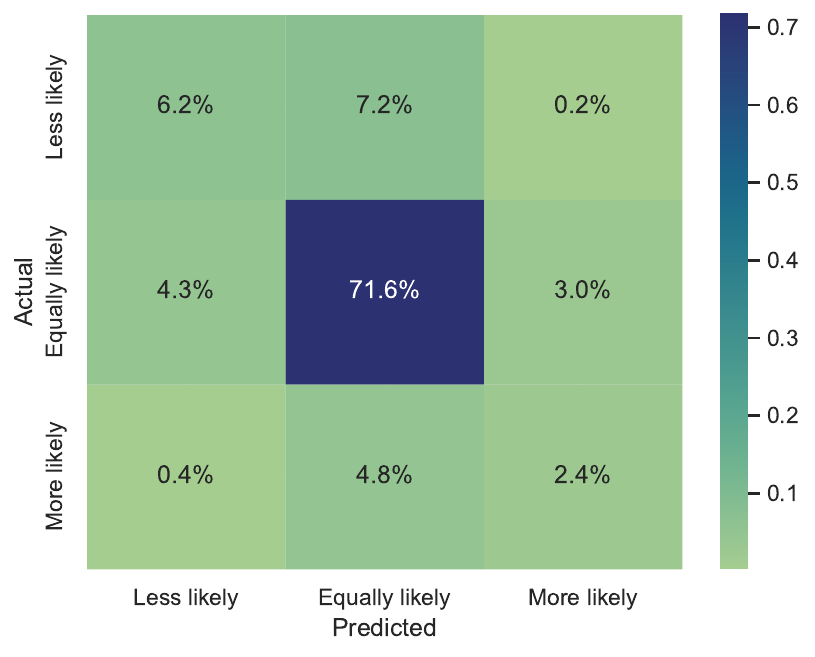}
    \caption{Confusion matrix for RoBERTa-\textsc{full}-17 with \textit{standard} evaluation} 
    \label{fig:3 - roberta-full standard eval}
    \end{subfigure}
\hspace{0.5cm}
    \begin{subfigure}[!htpb]{0.45\textwidth}
    \includegraphics[width=1\linewidth]{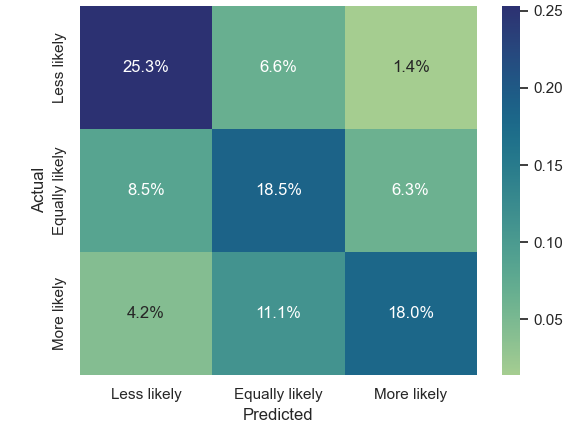}
    \caption{Confusion matrix for RoBERTa-\textsc{balanced}-17 with \textit{cross-balanced} evaluation} 
    \label{fig:4 - roberta-balanced cross-bal eval}
    \end{subfigure}
    \caption{Class-wise performance visualization
    }

\end{figure}

\begin{table}[!t]
\caption{Examples predictions based on RoBERTa-\textsc{full} and RoBERTa-\textsc{balanced} (dev set). We consider scenarios where (i) both approaches yield correct predictions for \textit{equally likely}, \textit{less likely}, and \textit{more likely}; (ii) \textit{equally likely} is misclassified as \textit{less likely} or \textit{more likely} by both models; and (iii) either methods captures instances of \textit{equally likely} that the other misses. For reasons of space, the presented sentence denotes the sentence pair consisting of two sentences, without and with the inserted adjectival modifier in \textless{}\textgreater{}~.}
\centering
\footnotesize
\begin{tabular}{@{}llll@{}}
\toprule
Scenario                 & Sentence with inserted adjectival modifier in \textless{}\textgreater{}~                                                                  & FUL & BAL \\ \midrule
equally likely          & A \textless{}coaxial\textgreater{}~laser is for light effects.                          & \Checkmark   & \Checkmark   \\
equally likely             & An \textless{}upright\textgreater{}~refrigerator is for making ice.           & \Checkmark   & \Checkmark    \\ \midrule
less likely          & An \textless{}errant\textgreater{}~student completes a course of study.                 & \Checkmark   & \Checkmark   \\
less  likely            & A \textless{}fake\textgreater{}~mug is for drinking tea.                                &\Checkmark   & \Checkmark   \\ \midrule
more likely                & The \textless{}effect\textgreater{}~of a great earthquake is a disaster.                & \Checkmark   & \Checkmark   \\
more likely             & A \textless{}neurological\textgreater{}~injury may cause bleeding.                      & \XSolidBrush   & \XSolidBrush   \\  \midrule
equally misc. less        & The \textless{}former\textgreater{}~speaker fields questions afterward.                 & \XSolidBrush   & \XSolidBrush     \\
equally misc. less         & \textless{}Suspicious\textgreater{}~land is for hunting.                                & \XSolidBrush   & \XSolidBrush   \\ \midrule
equally misc. more          & A \textless{}talented\textgreater{}~singer sings a ballad.                              & \XSolidBrush   & \XSolidBrush   \\ 
equally misc. more         & An \textless{}impressive\textgreater{}~gambler hopes to win money.                      & \XSolidBrush   & \XSolidBrush   \\ \midrule
\textsc{full}: misc. less, \textsc{bal}: equally  & \textless{}Wet\textgreater{}~rice is white.                      & \XSolidBrush   &  \Checkmark  \\
\textsc{full}: misc. more, \textsc{bal}: equally & An \textless{}international\textgreater{}~headliner stars in a movie. & \XSolidBrush   &  \Checkmark  \\ \midrule
\textsc{full}: equally, \textsc{bal}: misc. less  & A \textless{}potential\textgreater{}~wall is for containing plumbing.                    & \Checkmark  & \XSolidBrush    \\
\textsc{full}: equally, \textsc{bal}: misc. more & A \textless{}big\textgreater{}~kitchen is dirty.  & \Checkmark  & \XSolidBrush    \\
\bottomrule
\end{tabular}
\label{tab:examples}
\end{table}

\subsection{Model Error Analysis}

To investigate model errors, we visualize the performance of our best models, RoBERTa-\textsc{full} and RoBERTa-\textsc{balanced} fine-tuned with random seed 17, using confusion matrix heatmaps. Each cell shows the proportion of instances out of all dev set examples, based on \textit{standard} evaluation for RoBERTa-\textsc{full} and \textit{cross-balanced} evaluation for RoBERTa-\textsc{balanced}.
Similarly to Fig. 3 in \citet{emami-etal-2021-adept}, Fig.~\ref{fig:3 - roberta-full standard eval} shows that the model largely defaults to \textit{equally likely}, the most overrepresented class, resulting in poor performance on the remaining classes. Switching to \textit{cross-balanced} evaluation and the model trained on balanced data (Fig.~\ref{fig:4 - roberta-balanced cross-bal eval}) offers a more informative picture. RoBERTa-\textsc{balanced} proves most successful at identifying the class \textit{less likely}, while instances of \textit{equally likely} are frequently misclassified as either \textit{less likely} or \textit{more likely}, which has the effect of evening out performance across classes.

Further, we show example predictions in Table~\ref{tab:examples}. For reasons of space, the listed sentence includes the inserted adjectival modifier in brackets, while the actual sentence pair consists of two sentences. 
We first report examples that illustrate correctly classified examples, as well as examples where both models fail. This includes not capturing continuity in plausibility and failing to predict the majority class \textit{equally likely} correctly. Further, we include examples where RoBERTa-\textsc{full} does not catch the \textit{equally likely} class despite the the above-mentioned majority-class bias, whereas RoBERTa-\textsc{balanced} produces a correct prediction albeit being fine-tuned on a down-sampled subset. Finally, we present examples illustrating cases where RoBERTa-\textsc{balanced} fails to classify \textit{equally likely} correctly, which decreases overall performance. 

\section{Conclusion}

In this work, we tackled the task of predicting changes in semantic plausibility induced by a single adjectival modifier. Leveraging the \textsc{Adept} challenge dataset, we conducted an in-depth dataset suitability analysis. Our modeling experiments
reveal that both transformer models and sentence transformers struggle with the task at hand, particularly under a more realistic, balanced evaluation method. Notably, despite their conceptual alignment with the task, sentence transformers under-perform compared to models like RoBERTa and MPNet. In-depth error analyses and comparison to prior work explore potential sources for ST underperformance, and highlight the advantages and shortcomings of full and balanced train and test data. 
As directions for future work, we propose studying the effects of cross-balanced evaluation using model error analysis. We also recommend exploring a wider variety of (L)LMs and fine-tuning setups. Further, the \textsc{Adept} challenge dataset assumes several typological properties characteristic of English adjectival modification. For instance, sentence pairs are constructed by inserting a single adjective before a noun (\textit{dead horse}, prenominal adjective position), while, e.g. Romance languages often prefer positioning the adjective after the noun (French: \textit{cheval mort}, ``horse dead'').
Another relevant angle concerns the fact that adjectives in English do not account for gender, case, and number agreement, which is relevant in languages such as German, Russian, and Finnish. Studying such languages thus means addressing the additional challenge of making sure that models do not exploit agreement patterns but actually capture change in plausibility. 

\section*{Declarations}

The authors declare the use of AI assistants to improve grammar,  experiment with alternatives of formal notation, debug code underlying the ST model performance analysis, and format table boundaries.

\section*{Limitations}
 A first, obvious limitation of this work is that we study change in semantic plausibility exclusively on English data. We thus cannot be sure that our findings would generalize to other languages. 
Secondly, even though we address class imbalance by balancing out class distributions in the context of training data, future work could consider studying the effects of balancing train and test sets and the proposed cross-balancing evaluation through model error analysis. Furthermore, we did not systematically analyze potential effects of how instances are ordered within each class on cross-balanced evaluation. As the proposed approach takes consecutive cyclic slices of data from each class, instance ordering matters with regard to which examples end up evaluated together. Cross-balanced evaluation averages over shifts using a fixed trained model, reducing sensitivity, e.g. regarding the starting point; but it is still sensitive to how the dataset is originally ordered. In comparison, shuffling would makes cross-balanced evaluation behave more like repeated random sub-sampling of each class, instead of evaluating ordered consecutive slices of the dataset. 
Thirdly, we focused on transformer models as used by \citet{emami-etal-2021-adept} as well as MPNet in a standalone setting and as a pre-trained backbone model for our sentence transformer-based approach. With regard to the latter, we would recommend future work to further explore the hyperparameter setup during fine-tuning. 
Moreover, it is possible that with even larger or more powerful models, performance will improve, and existing model errors as well as the gap to human performance of assessments of change in event plausibility will shrink. 

\section*{Data Availability and Ethical Considerations}
\textsc{Adept} is released under a CC BY-SA 3.0 license and includes data from ConceptNet5, which is freely available under the Creative Commons Attribution-ShareAlike license (CC BY SA 4.0). Since the dataset is generated from English Wikipedia, Common Crawl and ConceptNet5, \textsc{Adept} might exhibit a range of biases which can be reproduced by language models. Following the dataset creators \citet{emami-etal-2021-adept}, we thus acknowledge that the dataset and any artifact using the data such as fine-tuned models can contain possible biases and should not be used to make deployment decisions or rule out failures. We also refer to the datasheet\footnote{\url{https://github.com/aemami1/adept/blob/master/DATASHEET_v2.md}} \citep{gebru2021datasheetsdatasets} provided by \citet{emami-etal-2021-adept}. Furthermore, our work aims to better understand and characterize language model capacities regarding event plausibility and how it changes when inserting a single adjectival modifier. Using open-source models only, we thus contribute to and highlight work underlining the importance of open access models and implementation.

\subsection*{Acknowledgments}
The authors are very grateful to Sabine Schulte im Walde for supervision, editorial advise, and valuable feedback on versions of this work. We thank Sinan Kurtyigit for discussion with regard to the formal notation of the cross-balanced evaluation approach. We further thank the guest editors and reviewers for their very constructive and nuanced feedback.

\subsection*{Funding}
The first and second authors have been supported by the ELLIS Unit Stuttgart. The third author received funding from the Hanns-Seidel-Stiftung's talent program. This study was further supported by the DFG Research Grant SCHU 2580–4 \textit{Multimodal Dimensions and Computational Applications of Abstractness}.

\subsection*{Competing interests}
The authors have no competing interests to declare. 

\begin{appendices}

\section{Details on Experimental Setup} \label{appsec:exp}

\paragraph{Transformers} To model semantic plausibility, we harness model implementations for \texttt{BERT-base-uncased}\footnote{\url{https://huggingface.co/google-bert/bert-base-uncased}} \citep{devlin-etal-2019-bert}, \texttt{DeBERTa-base}\footnote{\url{https://huggingface.co/microsoft/deberta-base}} \citep{deberta}, \texttt{RoBERTa-base}\footnote{ \url{https://huggingface.co/FacebookAI/roberta-base}} \citep{roberta} and \texttt{MPNet-base}\footnote{ \url{https://huggingface.co/microsoft/mpnet-base}} \citep{mpnet} provided through \texttt{huggingface}.

\paragraph{Sentence Transformers} 
We use the general-purpose sentence transformer model \texttt{all-mpnet-base-v2} provided through \texttt{huggingface}\footnote{\url{https://huggingface.co/sentence-transformers/all-mpnet-base-v2}} \citep{reimers-gurevych-2019-sentence}.

\end{appendices}




\end{document}